\documentclass[10pt]{article} 
\usepackage[accepted]{tmlr}

\usepackage{amsmath,amsfonts,bm}

\def\eqref#1{equation~\ref{#1}}

\def\1{\bm{1}}

\DeclareMathAlphabet{\mathsfit}{\encodingdefault}{\sfdefault}{m}{sl}
\SetMathAlphabet{\mathsfit}{bold}{\encodingdefault}{\sfdefault}{bx}{n}

\usepackage{latexsym}
\usepackage{hyperref}
\usepackage{url}

\usepackage{amsmath}
\usepackage{amssymb}
\usepackage{mathtools}
\usepackage{amsthm}

\usepackage{soul}
\usepackage{url}
\usepackage{graphicx}
\usepackage{booktabs}
\usepackage{algorithm}
\usepackage{array}
\usepackage{comment}
\usepackage{amsfonts}
\usepackage{multirow}
\usepackage{dblfloatfix}
\usepackage{multicol}
\usepackage{xcolor}
\usepackage{upgreek}
\usepackage{siunitx}
\usepackage{natbib}
\usepackage{xspace}
\usepackage{adjustbox}
\usepackage{wrapfig}
\usepackage{bm}
\usepackage{mathrsfs}
\usepackage{amsfonts}
\usepackage{amssymb}
\usepackage{color}
\usepackage{colortbl}
\usepackage{CJKutf8}
\usepackage{booktabs}
\usepackage{array} 
\usepackage{mwe}
\usepackage{titlesec}
\usepackage{enumitem}
\usepackage{microtype}
\usepackage{fancypar}
\usepackage{extarrows}
\usepackage{array}
\usepackage{nicefrac}       
\usepackage{supertabular,enumitem}
\usepackage{hyperref}

\usepackage{epsfig}
\usepackage{amssymb}
\usepackage{changepage}
\usepackage{enumerate}
\usepackage{amsfonts,lscape,color,url}
\usepackage{tabularx}
\usepackage{adjustbox}
\usepackage{overpic}
\usepackage{tablefootnote}

\definecolor{my_green}{RGB}{84,130,53}
\definecolor{my_red}{RGB}{192,2,0}

\newcolumntype{R}[1]{>{\raggedleft\arraybackslash}p{#1}}
\newcolumntype{L}[1]{>{\raggedright\arraybackslash}p{#1}}
\newcolumntype{C}[1]{>{\centering\arraybackslash}p{#1}}
\newcolumntype{A}[1]{>{\raggedleft\arraybackslash}p{#1}}

\let\classAND\AND
\let\AND\relax
\usepackage{algorithmic}

\let\AND\classAND
\AtBeginEnvironment{algorithmic}{\let\AND\algoAND}

\newcommand*\diff{\mathop{}\!\mathrm{d}}

\newcommand{\bs}[1]{\boldsymbol{#1}}
\newcommand{\ModelName}{\textsc{BDPL}}
\newcommand{\highlight}[1]{\textcolor{black}{#1}}

\newcommand{\blackbox}{black-box}
\newcommand{\whitebox}{white-box}

\definecolor{tablegray}{gray}{0.6}

\usepackage{pifont}
\newlength\savewidth\newcommand\shline{\noalign{\global\savewidth\arrayrulewidth
  \global\arrayrulewidth 1pt}\hline\noalign{\global\arrayrulewidth\savewidth}}

\makeatletter\renewcommand\paragraph{\@startsection{paragraph}{4}{\z@}
  {.5em \@plus1ex \@minus.2ex}{-.5em}{\normalfont\normalsize\bfseries}}\makeatother
\usepackage{xspace}

\theoremstyle{plain}

\theoremstyle{definition}

\theoremstyle{remark}

\title{Black-Box Prompt Learning for Pre-trained \\Language Models}

\author{\name Shizhe Diao \email sdiaoaa@connect.ust.hk \\
      \addr 
      The Hong Kong University of Science and Technology
      \AND
      \name Zhichao Huang \email zhuangbx@connect.ust.hk \\
      \addr 
      The Hong Kong University of Science and Technology
      \AND
      \name Ruijia Xu \email rxuaq@connect.ust.hk \\
      \addr 
      The Hong Kong University of Science and Technology
      \AND
      \name Xuechun Li \email xul021@ucsd.edu  \\
      \addr 
      University of California, San Diego
      \AND
      \name Yong Lin \email ylindf@connect.ust.hk \\
      \addr 
      The Hong Kong University of Science and Technology
      \AND
      \name Xiao Zhou \email xzhoubi@connect.ust.hk \\
      \addr 
      The Hong Kong University of Science and Technology
      \AND
      \name Tong Zhang\thanks{Joint with Google research} \email 	tongzhang@ust.hk \\
      \addr 
      The Hong Kong University of Science and Technology
      }

\def\month{02}  
\def\year{2023} 
\def\openreview{\url{https://openreview.net/forum?id=IvsGP7xRvm}} 

\begin{document}

\maketitle

\begin{abstract}
The increasing scale of general-purpose Pre-trained Language Models (\textbf{PLMs}) necessitates the study of more efficient adaptation across different downstream tasks. In this paper, we establish a Black-box Discrete Prompt Learning (\textbf{BDPL}) to resonate with pragmatic interactions between the cloud infrastructure and edge devices. 
Particularly, instead of fine-tuning the model in the cloud, we adapt PLMs by prompt learning, which efficiently optimizes only a few parameters of the discrete prompts.
Moreover, we consider the scenario that we do not have access to the parameters and gradients of the pre-trained models, except for its outputs given inputs. 
This black-box setting secures the cloud infrastructure from potential attack and misuse to cause a single-point failure, which is preferable to the white-box counterpart by current infrastructures. 
Under this black-box constraint, we apply a variance-reduced policy gradient algorithm to estimate the gradients of parameters in the categorical distribution of each discrete prompt. 
In light of our method, the user devices can efficiently tune their tasks by querying the PLMs bounded by a range of API calls. 
Our experiments on RoBERTa and GPT-3 demonstrate that the proposed algorithm achieves significant improvement on eight benchmarks in a cloud-device collaboration manner. 
Finally, we conduct in-depth case studies to comprehensively analyze our method in terms of various data sizes, prompt lengths, training budgets, optimization objectives, prompt transferability, and explanations of the learned prompts.\footnote{The code is available at \url{https://github.com/shizhediao/Black-Box-Prompt-Learning}.}
\end{abstract}

\setlength{\parskip}{1ex}

\section{Introduction}

Large Pre-trained Language Models (PLMs) have demonstrated impressive versatility across a wide spectrum of downstream tasks, via either fine-tuning (FT)~\citep{devlin2018bert, liu2019roberta, lewis2019bart, zhang2019dialogpt, yang2020styledgpt,diao2020zen,pan2022extremebert} or prompt-based learning (PL)~\citep{gao2020making, liu2021gpt, schick2021s, li2021prefix, liu2021pre}. Traditionally, these two tuning paradigms are conducted at a white-box setting, where the parameters and gradients are accessible since the model is usually open-sourced and can be duplicated in user devices. 
Despite the fact that white-box methods have made remarkable progress, however, the increasing scale of PLMs renders this setting implausible.
Nowadays, huge PLMs opt to serve users as commercial APIs deployed in the cloud, such as OpenAI GPT-3\footnote{\url{https://openai.com/api/}}. In particular, the service providers hide their model parameters and expose the query and prediction interface, which is termed the black-box setting in this paper. 

Although solving NLP problems with APIs in a black-box setting is considerably challenging, it is indeed aligned with the new norm of the current interplay between the cloud infrastructure and edge devices.
Specifically, from the position of cloud providers, it is reasonable to restrict the access of pre-trained model parameters since commercial, ethical, legal, security, and other concerns might be raised~\citep{bommasani2021opportunities}. 
First, under the white-box setting, the weaknesses and biases rooted in the underlying PLMs are at higher risk of being misused for harmful purposes. 
Second, the centralizing nature of PLMs exposes them to potential attacks to cause a single-point failure~\citep{krishna2019thieves}.
To this end, the black-box setting is more convincing to secure the cloud infrastructure from being condemned. As for the interests of user devices, the black-box paradigm grants them a more economical option. 
Otherwise, if we have access to the model's gradients, it requires transmitting gradients from cloud to device, causing high transmission costs.

\begin{table}[t]
\footnotesize
\centering
\def\w{20pt}
\begin{tabular}{l|ccccc}
\bf Methods & \bf Frozen & \bf Black-Box  & \bf Discrete & \bf Interpretable & \bf Learnable\\
\shline
Vanilla FineTuning & & & N/A & N/A & N/A \\
GPT-3's FineTuning\tablefootnote{\url{https://beta.openai.com/docs/guides/fine-tuning}\label{gpt3-finetune}} & \ding{51} & \ding{51} & N/A & N/A & N/A \\
FeatureProbe~\citep{peters2019tune} & \ding{51} & & & & \ding{51} \\
ManualPrompt & \ding{51} & \ding{51} & \ding{51} & \ding{51} & \\
InContextLearning\citep{NEURIPS2020_1457c0d6} & \ding{51} & \ding{51} & \ding{51} & \ding{51} & \\
PromptTuning~\citep{lester2021power} & \ding{51} & & & & \ding{51}\\
P-Tuning v2~\citep{DBLP:journals/corr/abs-2110-07602} & \ding{51} & & & & \ding{51}\\
AutoPrompt~\citep{shin2020autoprompt} & \ding{51} & & \ding{51} & \ding{51} & \ding{51}\\
BBT~\citep{sun2022black} & \ding{51} & \ding{51} & & & \ding{51}\\
{\ModelName} (ours) & \ding{51} & \ding{51} & \ding{51} & \ding{51} & \ding{51} \\
  \end{tabular}
\caption{
Comparison of different tuning methods.
\textbf{Frozen}: the pre-trained model is frozen and will not be updated.
\textbf{Black-Box}: there are no access to the parameters and gradients from the pre-trained model.
\textbf{Discrete}: the prompts are discrete tokens (compared with soft prompts).
\textbf{Interpretable}: the prompts are readable and interpretable.
\textbf{Learnable}: the prompts are parametric and learnable with explicit or estimated gradients (compared with manual prompts).
\textbf{N/A}: not applicable since the corresponding descriptions are for prompt learning. 
}
\label{tab:tuning_glossary}
\vspace{-1 em}
\end{table}

With this basic setting in mind, we further elaborate on a more pragmatic scenario, i.e., the discrete prompt learning under the black-box constraint. 
Particularly, we opt for the prompt learning mechanism instead of the fine-tuning counterpart, partially due to the fact that the prompt learning is more cost-effective by tuning fewer parameters and can eliminate the gap between pre-training and downstream transfer. 
Moreover, black-box fine-tuning\textsuperscript{\ref{gpt3-finetune}} requires users to upload their private labels and save the fine-tuned model on the server, which strongly relies on the cloud provider as the single-point trust for the security of private data and model. 
On the contrary, our black-box prompt learning allows users to store the private label and tune the prompt locally, preventing potential data leakage and protecting the users' commercial interests.
For instance, in our setting, each user device can query the output from the cloud and then update its prompts separately on its own data. 

It is noteworthy that we optimize discrete prompts, which are more interpretable to power users from different backgrounds to develop their own applications with PLMs. 
On the contrary, continuous prompt learning methods, e.g., BBT (black box tuning~\citep{sun2022black}), are difficult to interpret their learned prompts. 
Moreover, these methods fail to be directly applied to prediction APIs because APIs only accept discrete inputs (e.g., GPT-3). 
However, the discrete nature of our {\ModelName} allows commercial prediction APIs to directly take the learned prompt tokens without pain.
Overall, our established Black-Box Discrete Prompt Learning (BDPL) is closely in accordance with the recent progress of huge PLMs, whose comparisons with the existing settings can be found in table~\ref{tab:tuning_glossary}. 
As shown in the table, BDPL can be specified under the constraints that the PLMs are frozen and both their parameters and gradients are invisible and share the virtue of optimizing discrete, interpretable, and learnable prompt tokens simultaneously.

To embrace this paradigm shift of tuning PLMs, we design a policy gradient inspired framework that can be optimized without relying on the parameters and gradients of the pre-trained models. 
Specifically, we characterize the prompt learning procedure as a discrete token selection problem, where the proper prompt tokens are sampled according to a categorical distribution. 
Because the PLM's parameters are invisible and gradients cannot be back-propagated, the categorical distribution needs to be optimized by some gradient-free algorithms. 
We resort to the policy gradient algorithm to estimate the gradients without back-propagation. Moreover, to eliminate the high variance issue of policy gradient, we adopted a variance-reduced policy gradient estimator.

Experimental results on two kinds of datasets, \textit{i.e.}, datasets without domain-shift and datasets with domain-shift, demonstrate the effectiveness of the proposed {\blackbox} discrete prompt learning, which significantly improves the performance over a generic pre-trained model and outperforms all baseline models on eleven datasets.
The results confirm that incorporating {\blackbox} prompt learning for pre-trained models is an effective and efficient solution to the PLM adaptation.
We also present further analyses by investigating the effects of different training data sizes, prompt lengths, training budgets, and objectives. 
Our analyses demonstrated the robustness, scalability, and transferability of the proposed method.

The contributions of our work are as follows:

\noindent $\bullet$ We propose a new setting called {\blackbox}  prompt learning, where we only have access to the output of prediction APIs without the need to access the PLM's parameters or gradients.
The {\blackbox} prompt is optimized without the requirements of tuning pre-trained models, saving the fine-tuning costs. 

\noindent $\bullet$ We propose a new {\blackbox} discrete prompt learning (BDPL) method to solve this new problem, and demonstrate its effectiveness in dealing with domain shifts on various tasks. 

\noindent $\bullet$ 
We conduct comprehensive analyses on eleven benchmark datasets under cloud-device collaboration settings, demonstrating its effectiveness for commercial APIs.
{\ModelName} has a much wider range of applications than previous methods, such as transfer learning, model personalization, and decentralized training.

\begin{figure*}[t]
\begin{center}
\includegraphics[scale=0.48, trim=0 80 10 0,clip ]{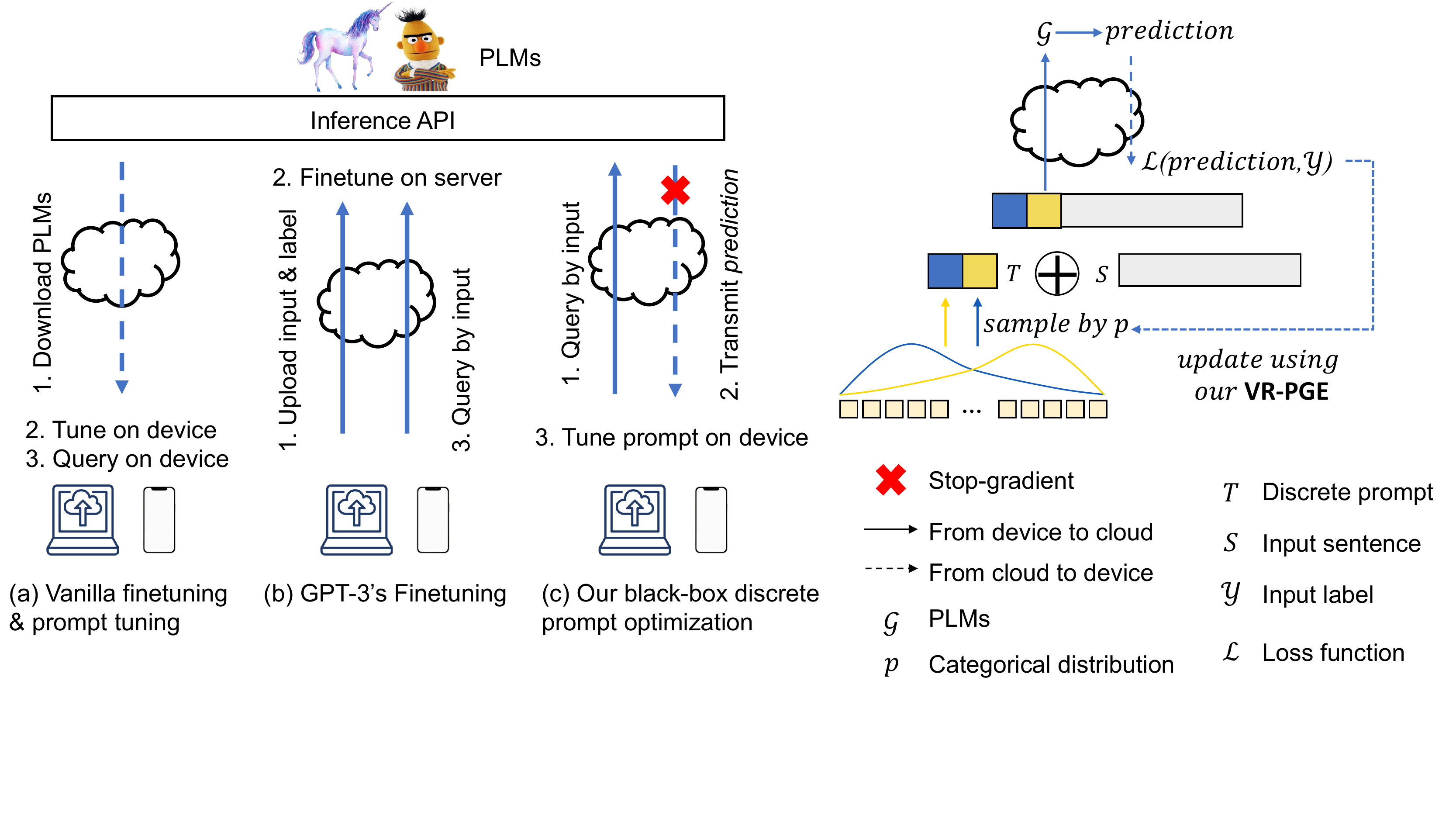}
\vskip -1 em
\caption{
Schematic illustrations of the comparisons across various tuning paradigms and the cloud-device interplay at the training phase of our algorithm. \textbf{Left:} (a) Vanilla finetuning and prompt tuning can be conducted on user devices in a white-box manner since the PLMs are feasible to be duplicated at user devices. After tuning, users can still access services of PLMs on the device. (b) Increasing scale hinders the democratizing of PLMs. In GPT-3's finetuning, users have to upload the input and associated labels to the server. After finetuning, the model is saved on the server. (c) In our black-box discrete prompt learning setting, users send queries to the server and then rely on PLMs' predictions to update their discrete prompts on the devices using a gradient-free optimization. \textbf{Right:} In our framework, the user query is created by concatenating the discrete prompt and the input sentence, where the prompt tokens are sampled from their categorical distributions respectively. After calculating the loss between the PLMs' predictions and input labels, we apply a variance-reduced policy gradient algorithm to estimate the gradient of the categorical distribution and update it accordingly.
}
\label{modelgraph}
\end{center}
\vskip -2 em
\end{figure*}

\section{Approach}

In our setting, the input is a sentence $S = {s_1 s_2 \cdots s_l \cdots s_m}$ with $s_l$ indicating the $l$-th token and the output corresponds to its category $y$.
Our goal is to learn $n$ discrete prompt tokens $T = t_1 t_2 \cdots t_i \cdots t_n = \mathcal{V}[j_1] \mathcal{V}[j_2] \cdots \mathcal{V}[j_i] \cdots \mathcal{V}[j_n]$, which are prepended to the input sentence to create the user query $[T, S]$. 
Note that $\mathcal{V}$ represents the vocabulary list consisting of a total of $N$ tokens, and $t_i = \mathcal{V}[j_i]$ is the $i$-th token in $T$ and $j_i$-th token in $\mathcal{V}$.

The overall architecture is shown in Figure~\ref{modelgraph}. 
During the {\blackbox} training, we freeze the prediction model $\mathcal{G}$ with a stop-gradient strategy, and only optimize the discrete prompts $T$. 
Here, we assume independent categorical distribution for each prompt index $j_i \sim \operatorname{Cat}(\boldsymbol{p}_i)$, where the random variable $j_i$ is sampled with the probability distribution $\boldsymbol{p}_i = [p_{i, 1}, \cdots, p_{i, N}]$ over the $N$ token indexes, where $\boldsymbol{p}_i \in \mathcal{C}$ and $\mathcal{C} = \{\boldsymbol{p}: \|\boldsymbol{p}\|_1 =1, 0 \preceq \bs{p} \preceq 1\}$. Since $\boldsymbol{p}_i$ is independent of each other, the joint probability of the whole discrete prompt is $P(T) = \Pi_{i=1}^{n} P(t_i) = \Pi_{i=1}^{n} p_{i, j_i}$.

Because the prediction model's parameters are invisible and gradients cannot be back-propagated to the prompts, it is no longer possible to directly update the prompts by back-propagating through $\nabla_{\boldsymbol{p}_i} \mathcal{L}(\mathcal{G}([T,S], y))$, where $y$ is the label.
Inspired by the policy gradient algorithm in discrete optimization, we resort to estimating the gradients without back-propagation to accomplish \textbf{black-box} training. 

For abbreviation, we denote the $\mathcal{L}(\mathcal{G}([T,S], y))$ as $\mathcal{L}(T)$ since $S, y$ can be deemed as constants here. 
By the virtue of the policy gradient estimator (PGE), we can optimize the loss function via forward propagation with:
\begin{equation}
    \mathbb{E}_{T}\left[\mathcal{L}(T)\right] = \int \mathcal{L}(T)P(T) \diff T,
\end{equation}
and estimate the gradient of $\boldsymbol{p}_i$ by:
\begin{equation}
    \begin{aligned}
     \nabla_{\boldsymbol{p}_i} \mathbb{E}_{T}\left[\mathcal{L}(T)\right]
    = &\int \mathcal{L}(T) \nabla_{\boldsymbol{p}_i} P(T) \diff T \\
    = &\int \mathcal{L}(T) \frac{P(T)}{P(T)} \nabla_{\boldsymbol{p}_i} P(T) \diff T \\
    = &\int P(T) \mathcal{L}(T) \nabla_{\boldsymbol{p}_i} \log P(T) \diff T \\
    = &\mathbb{E}_{P(T)}\left[\mathcal{L}(T)\nabla_{\boldsymbol{p}_i} \log \Pi_{j=1}^n P(t_j)\right] \\
    = &\mathbb{E}_{P(T)}\left[\mathcal{L}(T)\nabla_{\boldsymbol{p}_i} \log P(t_i)\right]
\end{aligned}
\end{equation}
The $j$-th component of $\nabla_{\boldsymbol{p}_i} \log P(t_i)$ could be solved explicitly by:
\begin{equation}
    \begin{aligned}
    & \nabla_{p_{i,j}} \log P(t_i) 
    =  \nabla_{p_{i,j}} \log p_{i, j_i}
    \label{eq:12}
\end{aligned}
\end{equation}
When $j = j_i$, it is obvious that $\nabla_{p_{i,j}} \log P(t_i) = \frac{1}{p_{i,j_i}}$. When $j \neq j_i$, equation (\ref{eq:12}) is calculated by:
\begin{equation}
    \begin{aligned}
     \nabla_{p_{i,j}} \log P(t_i) 
    = & \nabla_{p_{i,j}} \log (1 - \sum_{k=1, k \neq j_i}^{N} p_{i,k}) \\
    = & -\frac{1}{1 - \sum_{k=1, k \neq j_i}^{N} p_{i,k}} \\
    = & -\frac{1}{p_{i,j_i}}
\end{aligned}
\end{equation}
However, consistent with previous policy gradient applications~\citep{sutton1999policy,rezende2014stochastic, jang2016categorical, zhou2021efficient}, we observed that conventional PGE suffers from high variance, which makes it challenging to converge in practice.
Therefore, we adopted a variance-reduced policy gradient estimator (VR-PGE) as described in~\citet{williams1992simple,dong2020disarm,zhou2021efficient}. 
The estimated gradient is calculated by:

{\small
\begin{equation}
\boldsymbol{g}^{vr}_{\boldsymbol{p}_i}= \frac{1}{I-1} \sum_{k=1}^{I}\left(\mathcal{L}(T^{(k)}) -\frac{1}{I} \sum_{j=1}^{I} \mathcal{L}(T^{(j)})\right) \nabla_{\boldsymbol{p}_i}\log P(t_i) \label{eq:17} 
\end{equation}
}where $T^{(k)}, k = 1,\cdots, I$ are sampled independently from $P(T)$.

Thus, the prompt token distribution $\boldsymbol{p}_i$ can be updated by a projected stochastic gradient descent algorithm:
\begin{equation}
\boldsymbol{p}_{i} \gets \operatorname{proj}_{\mathcal{C}}(\boldsymbol{p}_{i}-\eta \cdot \boldsymbol{g}^{vr}_{\boldsymbol{p}_i}), i=1, \cdots, n
\label{eq:18}
 \end{equation}
where $\eta$ is the learning rate of prompt learning, $I$ is the sample size, and $\operatorname{proj}_{\mathcal{C}}$ is the projection calculation (details are presented in the Appendix).

\begin{algorithm}[t]
\caption{The {\blackbox} discrete optimization procedures.}
\label{alg:Framwork}
\begin{algorithmic}[1]
\REQUIRE
  {Input batch $S$,
  Label batch $Y$,
  Parameter of categorical distribution $\boldsymbol{p}_1, \cdots, \boldsymbol{p}_n$,
  Prediction model $\mathcal{G}$, Loss function $\mathcal{L}$.}
\FOR{$k \leq I$}
        \STATE {Sample $j_1^{(k)} \sim \operatorname{Cat}(\boldsymbol{p}_1), \cdots, j_n^{(k)} \sim \operatorname{Cat}(\boldsymbol{p}_n)$}
        \STATE $T^{(k)} = t_1^{(k)} \cdots t_n^{(k)} = \mathcal{V}[j_1^{(k)}]\cdots\mathcal{V}[j_n^{(k)}]$
\ENDFOR
\STATE $\mathcal{L}_{\text{avg}} = \frac{1}{I} \sum_{k=1}^{I} \mathcal{L}(\mathcal{G}[T^{(k)}, S], Y) $ 
\FOR{$i \leq n$}
\STATE{$
 \boldsymbol{g}^{vr}_{\boldsymbol{p}_i}= \frac{1}{I-1} \sum_{k=1}^{I} \nabla_{\boldsymbol{p}_i}\log P(t_i^{(k)})(\mathcal{L}(\mathcal{G}[T^{(k)}, S], Y) -\mathcal{L}_{\text{avg}})
$}
\STATE $\boldsymbol{p}_{i} \gets \operatorname{proj}_{\mathcal{C}}(\boldsymbol{p}_{i}-\eta \cdot  \boldsymbol{g}^{vr}_{\boldsymbol{p}_i})$
\ENDFOR
\STATE \textbf{return} $\boldsymbol{p}_{1}, \cdots \boldsymbol{p}_{n}$
\end{algorithmic}
\end{algorithm}

Here we introduce the detailed training procedure for updating the prompts using our proposed VR-PGE, whose mini-batch version is displayed in Algorithm~\ref{alg:Framwork}. 
Assuming the input data is divided into $B$ batches, and within each batch, we will perform $I$ iterations of sampling to reduce the variance of estimation. Specifically, at the $k$-th iteration within each batch, we first sample the sequence of prompt tokens $T^{(k)} = \mathcal{V}[j_{1}^{(k)}] \mathcal{V}[j_{2}^{(k)}] \cdots \mathcal{V}[j_{n}^{(k)}]$ according to the joint distribution $P(T)$. When $T^{(k)}$ is created, we will prepend it to the input sentence $S$ and feed the query $[T^{(k)}, S]$ into the black-box pre-trained language model $\mathcal{G}$, which will return back the prediction. In light of the model prediction and ground-truth label $Y$, we then calculate the loss $\mathcal{L}(\mathcal{G}[T^{(k)}, S], Y)$. Then the estimated gradients $\boldsymbol{g}^{vr}_{\boldsymbol{p}_i}$ for each $p_i$ could be obtained by executing Equation (\ref{eq:17}) after sampling all $I$ prompt sequences for the training batch. Finally, the categorical distributions are updated by Equation (\ref{eq:18}).

\paragraph{Vocabulary Construction}
A natural question is how to construct the vocabulary $V$ and what is the size $N$.
Inspired by the observation in~\citet{diao2021taming}, which revealed the importance of domain-specific and task-specific words and ngrams in representation learning, we introduce such important ngrams as prompt candidates.
Therefore, we adopt pointwise mutual information (PMI) to construct the vocabulary of candidate prompt tokens in an unsupervised way.
For each sentence in the training set, we calculate the PMI by
\begin{equation}
\begin{aligned}
    \text{PMI}(\bar{x}, \widetilde{x}) = \log \frac{p(\bar{x}\widetilde{x})}{p(\bar{x})p(\widetilde{x})},
\end{aligned}
\end{equation}
where $\bar{x}$ and $\widetilde{x}$ are two adjacent words in the sentence, and $p(x)$ is the probability of an n-gram $x$.
If the PMI score between these two adjacent words is high, they have a high probability of co-occurrence and are more likely to form an n-gram, suggesting they are good collocation pairs.
If the PMI score is lower than a threshold $\sigma$, a delimiter is inserted between $\bar{x}$ and $\widetilde{x}$.
As a result, the sentence will be segmented by several delimiters.
Finally, we obtain a list of ngrams $V$ by extracting those consecutive words after segmentation and with a frequency of at least $f$.
As for the size $N$, we observe that large $N$ will cause an unstable optimization process and even divergence.
Therefore, we keep $N$ between 50 and 200.

\section{Experimental Settings}
In this section, we first introduce the datasets and evaluation metrics (\S \ref{sec: Datasets}), followed by the baseline models (\S \ref{sec: Baselines}). 
Lastly, we describe the implementation details (\S \ref{sec: Implementation}).

\subsection{Datasets and Evaluation Metrics}
\label{sec: Datasets}
\highlight{In order to examine the model's ability in generic classification tasks as well as domain-specific classification tasks, we include seven datasets from the GLUE benchmark~\citep{wang2018glue}: MNLI~\citep{williams2018broad}, QQP~\citep{iyer2017first}, SST-2~\citep{socher2013recursive},} MRPC~\citep{dolan2005automatically},\textsc{ CoLA}~\citep{warstadt2019neural}, QNLI~\citep{wang2018glue}, RTE~\citep{dagan2005pascal,haim2006second,giampiccolo2007third,bentivogli2009fifth}, and four domain-specific datasets: \textsc{CitationIntent}~\citep{citation_intent}, \textsc{SciERC}~\citep{sciie}, RCT~\citep{rct}, \textsc{HyperPartisan}~\citep{hyp} from specific domains including computer science, biomedical science and news following~\citet{gururangan2020don,diao2021taming}.
The statistics of these datasets are shown in Table \ref{table:data_statistics}.
Considering the data sparsity issue and large query costs\footnote{For example, only one training epoch on 10, 000 sentences with 300, 000 tokens will cause 6 USD costs for GPT-3-Davinci. Not to mention tens of hundreds of rounds of training.} in cloud-device collaboration, we conduct our experiments on a popular and more realistic setting --- few-shot learning, where huge models have shown their powerful ability~\citep{NEURIPS2020_1457c0d6}.
We follow~\citet{perez2021true} to simulate a true $k$-shot learning setting.
We randomly sample $k$ data from the original training set for each class to construct the training set and another different $k$ data to construct the validation set.
The original validation set will be used as the test set.
Because the size of the QQP and RCT validation sets is too large, we randomly sample 1K data to save costs. 
We adopt Matthews Correlation Coefficient for \textsc{CoLA}, F1-score for QQP, MRPC, \textsc{CitationIntent}, \textsc{SciERC}, \textsc{HyperPartisan}, RCT, and accuracy for SST-2, RTE, QNLI, SST-2, IMDB, CR, MR, and MPQA following~\citet{wang2018glue, diao2021taming}.
\highlight{MNLI results are an average of MNLI-match and MNLI-mismatch accuracy.}

\begin{table*}[t]
\normalsize
\centering
\begin{tabular}{cccccccc}
\toprule
\textbf{Dataset} & $\mathbf{|L|}$ & $|$\textbf{Train}$|$ & $|$\textbf{Dev}$|$ & $|$\textbf{Test}$|$ & \textbf{Type} & \textbf{Metrics} & \textbf{Domain} \\
\midrule
\multicolumn{8}{c}{\textit{Generic Tasks}}
\\\midrule
\highlight{MNLI} & \highlight{3} & \highlight{393K} & \highlight{9.8K} & \highlight{9.8K} & \highlight{NLI} & \highlight{acc.} & \highlight{fiction, reports} \\
\highlight{QQP} & \highlight{2} & \highlight{364K} & \highlight{40K} & \highlight{391K} & \highlight{paraphrase} & \highlight{F1} & \highlight{Quora} \\
\highlight{SST-2} & \highlight{2} & \highlight{6.7K} & \highlight{872} & \highlight{1.8K} & \highlight{sentiment} & \highlight{acc.} & \highlight{movie reviews} \\
MRPC & 2 & 3.7K & 408 & 1.7K & paraphrase & F1 & news \\
CoLA & 2 & 8.6K & 1K & 1K & acceptability &  Matthews corr. & books, articles \\
QNLI & 2 & 105K & 5.5K & 5.5K & NLI &  acc. & Wikipedia \\
RTE & 2 & 2.5K & 277 & 3K & NLI & acc. & news, Wikipedia \\
\midrule
\multicolumn{8}{c}{\textit{Domain-Specific Tasks}}
\\\midrule
CI & 6 & 1.6K & 114 & 139 & citation intent & F1 & computer science \\
SE & 7 & 3.2K & 455 & 974 & relation classification & F1 & computer science \\
RCT & 5 & 180K & 30K & 30K & abstract sentence roles & F1 & biomedical \\
HP & 2 & 516 & 64 & 65 & review helpfulness & F1 & reviews \\
\bottomrule
\end{tabular}
\caption{The statistics of seven datasets in the generic domain and four datasets in the specific domain.
CI, SE, HP denote \textsc{CitationIntent}, \textsc{SciERC}, \textsc{HyperPartisan}, respectively.
$\mathbf{|L|}$: number of classes for classification tasks.
Note that we sample the few-shot training split and development split from the original training split for few-shot setting as described in Section~\ref{sec: Datasets}.
}
\label{table:data_statistics}
\vskip -1 em
\end{table*}

\subsection{Baselines}
\label{sec: Baselines}

For GPT-3-based models, because previous white-box tuning methods and black-box continuous prompt tuning methods (e.g., BBT) cannot be applied to GPT-3, we compare our model with the following baselines.
\begin{itemize}[leftmargin=*,label=$\bullet$,noitemsep,partopsep=0pt,topsep=0pt,parsep=0pt]
\item \textbf{GPT-3's FineTuning}\footnote{\url{https://beta.openai.com/docs/guides/fine-tuning}}: a GPT-3 inference API that is fine-tuned entirely on a labeled dataset (black-box).
\item \textbf{{ManualPrompt}}: a GPT-3 inference API with manually composed prompts to conduct the zero-shot evaluation. 
The human-written prompts are shown in Appendix~\ref{appendix:manual_templates} (black-box).
\item \textbf{InContextLearning}~\citep{NEURIPS2020_1457c0d6}: a GPT-3 inference API with a set of examples containing training sentences and labels as the input prefix, which is then prepended to the input texts (black-box).
\end{itemize}

For RoBERTa-based models, we adopt the RoBERTa-large as the backbone and the following tuning methods. 
\begin{itemize}[leftmargin=*,label=$\bullet$,noitemsep,partopsep=0pt,topsep=0pt,parsep=0pt]
\item \textbf{Vanilla FineTuning}~\citep{liu2019roberta}: a RoBERTa-large model that is fine-tuned entirely on a labeled dataset (white-box).
\item \textbf{PromptTuning}~\citep{lester2021power}: a frozen RoBERTa-large model with continuous prompt embeddings prepended to the input, and learned by gradients (white-box).
\item \textbf{P-Tuning v2}~\citep{DBLP:journals/corr/abs-2110-07602}: a frozen RoBERTa-large model with continuous prompt embeddings prepended to each layer, and learned by gradients (white-box).
\item \textbf{AutoPrompt}~\citep{shin2020autoprompt}: 
a frozen RoBERTa-large model with discrete prompts optimized based on gradient-guided search (white-box).
\item \textbf{FeatureProbe}~\citep{peters2019tune}: a frozen RoBERTa-large model outputs the features given inputs and a newly added classification layer is trained with the gradients (white-box).
\item \textbf{ManualPrompt}: a frozen RoBERTa-large model with manually composed prompts to conduct the zero-shot evaluation. 
The human-written prompts are shown in Appendix~\ref{appendix:manual_templates} (black-box).
\item \textbf{InContextLearning}~\citep{NEURIPS2020_1457c0d6}: a frozen RoBERTa-large model with a set of examples containing training sentences and labels as the input prefix, which is then prepended to the input texts (black-box).
\item \textbf{\textsc{BBT}}~\citep{sun2022black}: a frozen RoBERTa-large model with continuous prompts that are optimized by covariance matrix adaptation evolution strategy (black-box).
\item \highlight{\textbf{RLPrompt}~\citep{deng2022rlprompt}: a frozen RoBERTa-large model with discrete prompts that are generated by a policy network and optimized by a reward function (black-box).}
\end{itemize}

\subsection{Implementation}
\label{sec: Implementation}
For GPT-3 experiments, we conduct experiments with four variants: GPT-3-Ada, GPT-3-Babbage, GPT-3-Curie, and GPT-3-Davinci.
The batch size of training and evaluation is set to 4 to fulfill the query length limit (i.e., 2048).
We call the APIs directly from OpenAI's services\footnote{\url{https://openai.com/api/}}.
For RoBERTa-large experiments, we initialize it with pre-trained weights by Huggingface’s Transformers library\footnote{\url{https://github.com/huggingface/transformers}}. 
The batch size of training and evaluation is set to 16 and 32, respectively.
The number of API calls is limited to 8000 across all datasets.

For {\ModelName}, we optimize the prompts by AdamW~\citep{loshchilov2018decoupled} for 30 epochs with a learning rate of $1\times 10^{-4}$. 
The prompt length is 50, and the size of the candidate prompt list $N$ is 100.
Other hyper-parameters are detailed in the Appendix~\ref{appendix:hyper-params}.

\begin{table}[t]
\centering
\setlength{\tabcolsep}{0.9mm}{
\begin{tabular}{l|ccccccccccc|c|c}
\toprule
\textsc{Dataset} & MNLI & QQP & SST-2 & \textsc{MRPC} & \textsc{CoLA} & \textsc{QNLI} & \textsc{RTE}  & \textsc{CI} & \textsc{SE} & \textsc{RCT} & \textsc{HP} & \textsc{Avg.} & \textsc{\$Cost}\\
\midrule
 \multicolumn{14}{c}{\textit{GPT-3 Ada}}
\\ \midrule
FT & 38.5\tiny{0.8} & 44.5\tiny{1.4} & 71.6\tiny{1.2} & 45.7\tiny{4.3} & 0.0\tiny{0.0} & 49.8\tiny{2.1} & 52.7\tiny{1.2} & 27.7\tiny{3.2} & 3.5\tiny{0.3} & 57.0\tiny{4.2} & 24.4\tiny{1.4} & 37.8 & 5.6 \\
MP & 26.5\tiny{0.9} & 31.2\tiny{1.8} & 63.1\tiny{1.3} & 35.6\tiny{2.1} & 0.0\tiny{0.0} & 45.6\tiny{1.5} & 47.3\tiny{2.0} & 26.9\tiny{2.0} & 1.2\tiny{0.6} & 15.8\tiny{2.4} & 15.2\tiny{1.8} & 28.0 & 0.5 \\
{ICL} & 36.3\tiny{0.7} & 40.3\tiny{1.3} & 64.6\tiny{0.8} & 40.5\tiny{1.5} & 1.3\tiny{1.4} & 48.8\tiny{1.7} & 48.7\tiny{1.3} & 28.2\tiny{0.3} & 2.5\tiny{0.4} & 22.7\tiny{2.6} & 20.0\tiny{2.3} & 32.2 & 5.7 \\
\midrule
{\ModelName} & 37.1\tiny{0.7} & 45.1\tiny{0.5} & 68.8\tiny{0.9} & 43.2\tiny{2.5} & 2.0\tiny{0.4} & 51.2\tiny{0.4} & 52.7\tiny{1.3} & 28.3\tiny{1.5} & 3.8\tiny{0.3} & 45.7\tiny{2.9} & 22.4\tiny{1.7} & 36.4 & 3.2 \\
\midrule
 \multicolumn{14}{c}{\textit{GPT-3 Babbage}}
\\ \midrule
FT  & 40.7\tiny{0.5} & 46.2\tiny{1.4} & 87.4\tiny{1.5} & 66.4\tiny{1.7} & 0.3\tiny{0.1} & 50.9\tiny{0.2} & 52.3\tiny{1.0} & 5.2\tiny{0.4} & 4.1\tiny{1.0} & 61.1\tiny{5.2} & 33.3\tiny{1.3} & 40.7 & 8.5 \\
MP & 28.9\tiny{0.8} & 34.1\tiny{1.2} & 83.5\tiny{1.2} & 62.4\tiny{3.2} & 0.2\tiny{0.1} & 48.8\tiny{1.4} & 51.2\tiny{0.6} & 31.4\tiny{2.8} & 1.7\tiny{0.5} & 21.7\tiny{2.3} & 27.2\tiny{1.5} & 35.6 & 0.6 \\
{ICL} & 35.7\tiny{0.9} & 45.2\tiny{1.9} & 86.2\tiny{1.4} & 65.4\tiny{1.7} & 2.6\tiny{0.0} & 48.3\tiny{0.9} & 51.5\tiny{0.4} & 13.1\tiny{1.5} & 2.5\tiny{0.9} & 36.7\tiny{1.8} & 32.2\tiny{1.4} & 38.1 & 7.1 \\
\midrule
{\ModelName}  & 41.0\tiny{0.6} & 50.4\tiny{1.5} & 86.4\tiny{1.1} & 67.7\tiny{1.2} & 2.8\tiny{0.1} & 52.1\tiny{0.3} & 53.1\tiny{1.0} & 40.2\tiny{2.5} & 3.2\tiny{0.8} & 45.2\tiny{2.2} & 30.4\tiny{2.3} & 43.0 & 4.0 \\
\midrule
 \multicolumn{14}{c}{\textit{GPT-3 Curie}}
\\ \midrule
FT & 42.2\tiny{2.8} & 53.3\tiny{1.4} & 88.9\tiny{3.1} &  76.3\tiny{2.1} & 3.4\tiny{1.3} & 49.0\tiny{1.3} & 54.5\tiny{1.7} & 28.4\tiny{1.9} & 5.1\tiny{0.8} & 50.6\tiny{1.3} & 43.3\tiny{1.5} & 45.0 & 42.3 \\
MP & 34.5\tiny{1.9} & 44.3\tiny{2.1} & 84.2\tiny{1.4} & 73.3\tiny{1.2} & 2.0\tiny{0.6} & 47.2\tiny{0.9} & 44.0\tiny{1.2} & 19.2\tiny{1.3} & 2.8\tiny{0.3} & 31.0\tiny{1.4} & 37.1\tiny{1.5} & 38.1 & 2.5 \\
{ICL} & 38.0\tiny{2.1} & 47.2\tiny{2.3} & 87.0\tiny{1.9} & 81.0\tiny{1.3} & 2.8\tiny{0.0} & 46.8\tiny{1.1} & 46.2\tiny{1.6} & 15.2\tiny{1.9} & 4.8\tiny{1.3} & 50.1\tiny{2.3} & 39.0\tiny{2.3} & 41.6 & 28.5 \\
\midrule
{\ModelName} & 42.5\tiny{1.9} & 52.0\tiny{1.5} & 88.0\tiny{2.2} & 82.6\tiny{1.1} & 4.0\tiny{1.3} & 50.1\tiny{0.8} & 55.8\tiny{1.3} & 25.5\tiny{1.8} & 3.4\tiny{1.8} & 49.6\tiny{2.4} & 39.4\tiny{1.6} & 44.8 & 16.2 \\
\midrule
 \multicolumn{14}{c}{\textit{GPT-3 Davinci}}
\\ \midrule
FT & 60.2\tiny{3.8} & 67.8\tiny{2.1} & 92.9\tiny{2.4} & 84.6\tiny{1.3} & 55.3\tiny{1.5} & 54.2\tiny{2.5} & 57.0\tiny{1.2} & 35.4\tiny{1.7} & 10.3\tiny{2.3} & 51.6\tiny{2.7} & 60.1\tiny{1.8} & 57.2 & 423.2 \\
MP & 40.2\tiny{2.5} & 39.2\tiny{1.6} & 86.7\tiny{2.7} & 69.7\tiny{2.1} & 55.2\tiny{2.4} & 28.0\tiny{1.3} & 55.3\tiny{1.9} & 25.6\tiny{2.0} & 4.9\tiny{1.0} & 26.8\tiny{1.9} & 52.6\tiny{1.4} & 44.0 & 25.0 \\
{ICL} & 52.7\tiny{2.9} & 55.6\tiny{3.4} & 87.2\tiny{3.3} & 82.4\tiny{1.7} & 56.7\tiny{2.0} & 17.9\tiny{1.7} & 56.6\tiny{2.3} & 30.1\tiny{3.0} & 9.2\tiny{1.5} & 44.4\tiny{2.2} & 55.4\tiny{2.5} & 49.8 & 206.1 \\
\midrule
{\ModelName} & 54.6\tiny{2.4} & 57.8\tiny{2.1} & 89.3\tiny{3.0} & 83.4\tiny{1.4} & 58.4\tiny{1.4} & 56.2\tiny{1.5} & 57.2\tiny{0.8} & 34.6\tiny{2.0} & 6.6\tiny{2.1} & 48.8\tiny{2.5} & 58.5\tiny{2.4} & 55.0 & 161.5 \\
\bottomrule
\end{tabular}
}
\vskip -1 em
\caption{
The overall performance of {\blackbox} prompt and the comparison on eleven datasets with GPT-3.
We report average scores across three random seeds, with standard deviations as subscripts.
\textsc{Avg.} denotes the average score across all tasks.
\textsc{\$Cost} denotes the money cost in US Dollar for calling GPT-3's API during training and inference.
FT: GPT-3's FineTuning. 
MP: ManualPrompt. 
ICL: InContextLearning. 
}
\vskip -1 em
\label{tab:overall_performance_gpt}
\end{table}

\section{Experimental Results}
\label{sec:performance}
The overall results on eleven datasets are reported in Tables \ref{tab:overall_performance_gpt} and \ref{tab:overall_performance_roberta}.

We first verify our proposed method's effectiveness on a purely black-box setting with GPT-3 APIs.
From Table~\ref{tab:overall_performance_gpt}, {\ModelName} shows great superiority across eleven datasets.
Compared with ManualPrompt and InContextLearning, {\ModelName} demonstrates significant improvements, which are 8.35\% and 4.35\% on average of Ada, Babbage, Curie, and Davinci models.
{\ModelName} also achieves comparable performance with GPT-3's fine-tuning, which requires large money costs and uploading user's data. 
In Babbage, {\ModelName} even outperforms GPT-3's fine-tuning. 
As the experiments are conducted on the few-shot setting, where a small number of data are available to fine-tune the models' parameters, for large models like GPT-3, overfitting could be a serious problem that deteriorates the performance so that fine-tuning is inferior to {\ModelName}. 
Although careful adjustment of the fine-tuning algorithm may mitigate overfitting and improve accuracy, it needs a lot of manual effort and money costs, which is implausible for cloud-device collaboration.
Moreover, it is observed that ManualPrompt and InContextLearning with less capable versions of GPT-3 (e.g., Ada and Babbage) fail in some challenging datasets (e.g., \textsc{CoLA} and SE) but {\ModelName} performs well on them.
With the increase of the model capacity (from Ada to Davinci), we observed ManualPrompt and InContextLearning could also solve them, which is consistent with recent observations of large model's emergent abilities~\citep{wei2022emergent}.
From this perspective, our method offers another option in addition to increasing model size, which is an efficient solution for less capable models to perform challenging tasks.

In addition to the auto-regressive model, we also conduct additional experiments with an encode-only model, RoBERTa-large.
Because the weights of RoBERTa-large are released and gradients can be leveraged, several white-box baseline models are introduced for comparison.
First, our model outperforms all black-box methods, demonstrating the effectiveness of our proposed black-box discrete prompt optimization.
Second, {\ModelName} achieves comparable performance compared with white-box prompt-based methods including both discrete and continuous prompts.
\highlight{It is observed that BDPL even outperforms some white-box methods (e.g., PromptTuning and AutoPrompt).
We attribute this phenomenon to the overfitting of white-box methods in terms of the given few-shot examples while BDPL does not suffer severe overfitting due to its exploration mechanism.
We perform an ablation study in Section~\ref{appendix:overfitting} to reveal the effect of data size and accuracy.}
Given that white-box prompt tuning methods cannot be applied in black-box settings when the gradients are unavailable, previous black-box methods such as InContextLearning and BBT can achieve 2.82\% and 6.66\% improvement on average over ManualPrompt.
\highlight{{\ModelName} outperforms the previous black-box methods BBT and RLPrompt by an average of 2.7\% and 2.5\%, respectively.}
Compared with BBT, our method, {\ModelName}, not only outperforms it but is also more practical considering its discrete nature. 
BBT is optimizing continuous prompts and cannot be directly fed into current prediction APIs.
We also notice that there is still a large gap between FineTuning and all other methods.
FineTuning updates the full model with gradients and huge parameters, serving as an upper bound for all methods.
Across eleven tasks, it is observed that the {\ModelName} on domain-specific datasets is as effective as on generic datasets. 
While it is known that domain shift introduced difficulty for models to deal with, {\ModelName} offers an effective solution to domain-specific datasets.

\begin{table}[t]
\centering
\setlength{\tabcolsep}{1mm}{
\begin{tabular}{l|ccccccccccc|r}
\toprule
\textsc{Dataset} & MNLI & QQP & SST-2 & \textsc{MRPC} & \textsc{CoLA} & \textsc{QNLI} & \textsc{RTE}  & \textsc{CI} & \textsc{SE} & \textsc{RCT} & \textsc{HP} & \textsc{Avg.} \\
\midrule
 \multicolumn{13}{c}{\textit{White-Box Methods}}
\\ \midrule
FT & 50.8\tiny{1.2} & 60.8\tiny{1.9} & 86.5\tiny{2.0} & 78.4\tiny{1.3} & 20.4\tiny{1.9} & 53.2\tiny{1.8} & 55.6\tiny{2.5} & 37.4\tiny{1.7} & 23.1\tiny{1.6} & 45.2\tiny{5.2} & 55.5\tiny{2.3} & 51.5 \\
{PromptTuning} & 36.5\tiny{0.9} & 50.2\tiny{1.5} & 70.7\tiny{2.6} & 52.7\tiny{3.4} & 8.0\tiny{0.7} & 53.5\tiny{1.6} & 56.3\tiny{1.6} & 34.4\tiny{2.6} & 28.6\tiny{2.5} & 36.7\tiny{3.1} & 47.4\tiny{3.2} & 43.2 \\
{P-Tuning v2} & 44.2\tiny{1.7} & 57.4\tiny{2.4} & 80.4\tiny{1.2} & 62.4\tiny{2.0} & 8.9\tiny{2.7} & 51.5\tiny{1.3} & 53.1\tiny{1.7} & 31.4\tiny{4.2} & 24.6\tiny{2.5} & 35.4\tiny{3.9} & 55.4\tiny{4.2} & 45.9 \\
{AutoPrompt} & 40.1\tiny{1.5} & 45.7\tiny{1.3} & 71.5\tiny{2.1} & 63.8\tiny{3.1} & 5.4\tiny{2.3} & 50.2\tiny{1.3} & 52.1\tiny{1.6} & 27.9\tiny{2.9} & 21.5\tiny{2.5} & 29.6\tiny{2.5} & 40.6\tiny{3.8} & 40.8 \\
{FeatureProbe} & 46.5\tiny{1.8} & 56.3\tiny{1.1} & 79.5\tiny{1.6} & 68.9\tiny{1.7} & 15.6\tiny{1.2} & 50.5\tiny{0.2} & 54.1\tiny{2.5} & 22.3\tiny{2.0} & 20.8\tiny{3.6} & 31.2\tiny{4.7} & 60.1\tiny{2.6} & 46.0 \\
\midrule
 \multicolumn{13}{c}{\textit{Black-Box Methods}}
\\ \midrule
{ManualPrompt} & 35.9\tiny{1.3} & 49.8\tiny{0.9} & 77.2\tiny{2.1} & 70.4\tiny{1.6} & 0.6\tiny{0.0} & 49.2\tiny{1.1} & 48.2\tiny{0.6} & 12.3\tiny{2.4} & 9.6\tiny{1.4} & 11.7\tiny{1.5} & 35.7\tiny{1.6} & 36.4 \\
ICT & 37.2\tiny{1.6} & 50.1\tiny{0.9} & 82.8\tiny{2.1} & 72.1\tiny{2.3} & 1.1\tiny{0.4} & 50.8\tiny{0.5} & 49.3\tiny{2.3} & 14.6\tiny{1.7} & 9.2\tiny{1.5} & 25.8\tiny{1.6} & 38.5\tiny{2.4} & 39.2 \\
{BBT} & 40.6\tiny{2.5} & 55.2\tiny{3.1} & 85.3\tiny{3.9} & 66.4\tiny{3.7} & 5.5\tiny{2.7} & 55.4\tiny{3.2} & 52.6\tiny{2.2} & 17.4\tiny{5.4} & 16.4\tiny{0.9} & 31.7\tiny{1.5} & 47.2\tiny{4.8} & 43.1 \\
\highlight{RLPrompt} & \highlight{42.8\tiny{3.2}} & \highlight{53.7\tiny{2.2}} & \highlight{88.4\tiny{1.9}} & \highlight{68.9\tiny{2.1}} & \highlight{5.0\tiny{1.1}} & \highlight{52.6\tiny{1.4}} & \highlight{51.8\tiny{1.8}} & \highlight{19.2\tiny{3.3}} & \highlight{18.8\tiny{1.5}} & \highlight{30.1\tiny{2.7}} & \highlight{44.9\tiny{2.4}} & \highlight{43.3} \\
\midrule
{\ModelName} & 42.5\tiny{1.8} & 56.4\tiny{1.9} & 87.6\tiny{2.1} & 78.1\tiny{3.7} & 4.6\tiny{1.2} & 53.1\tiny{1.1} & 53.5\tiny{0.9} & 24.0\tiny{1.3} & 21.5\tiny{2.0} & 36.6\tiny{3.2} & 45.6\tiny{3.4} & 45.8 \\
\bottomrule
\end{tabular}
}
\vskip -0.5 em
\caption{
The overall performance of {\blackbox} prompt and the comparison on eleven datasets with RoBERTa-large.
We report average scores across three random seeds, with standard deviations as subscripts.
\textsc{Avg.} denotes the average score across all tasks.
FT: Vanilla FineTuning.
ICT: InContextLearning.
}
\label{tab:overall_performance_roberta}
\vskip -1 em
\end{table}

\section{Analysis}
We analyze several aspects of {\ModelName}, including the effects of different training data sizes, prompt lengths, training budgets, and learning objectives.
In addition, we examine the transferability of our learned prompts under the transfer learning setting and the explanation of prompts. 
We choose GPT-3 Babbage as the backbone model in the following discussion.
The details are illustrated in this section.

\subsection{Ablation Study}
\begin{figure*}[t]
\begin{center}
\includegraphics[scale=0.49, trim=0 100 0 0,clip ]{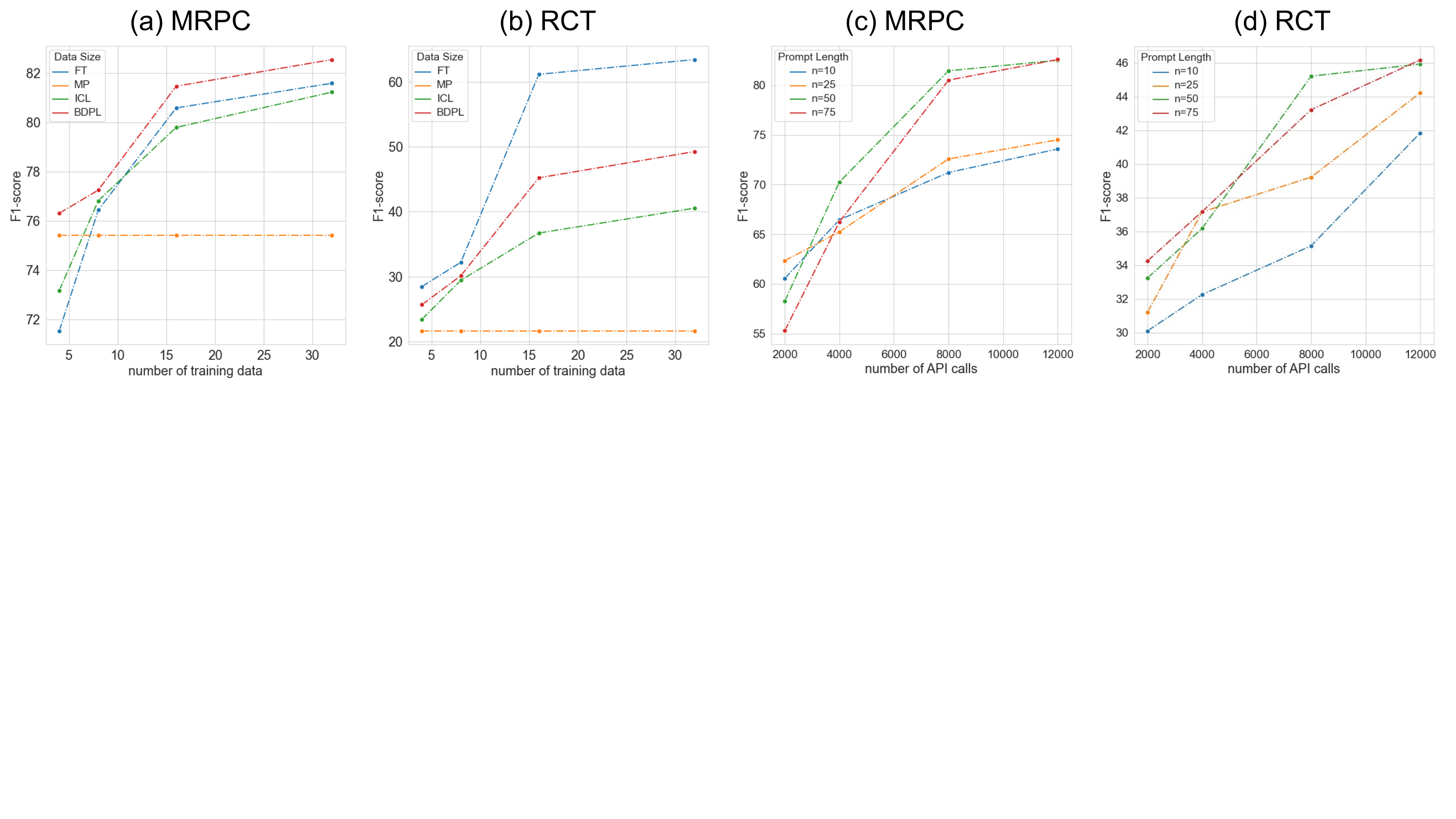}
\vskip -10 em
\caption{The effects of training data size, prompt length, and the number of API calls on MRPC and RCT.
FT, MP, and ICL denote GPT-3's FineTuning, ManualPrompt, and InContextLearning, respectively.
}
\vskip -1 em
\label{fig:ablation_study}
\end{center}
\end{figure*}

\paragraph{Effects of Training Data Size}
\label{Effects of Train Data Size}
First, we analyze the effects brought by four different training data sizes: 4-shot, 8-shot, 16-shot, and 32-shot.
Experiments are conducted on MRPC and RCT datasets.
As shown in the left part of Figure \ref{fig:ablation_study} (a) and (b), with the increase in training data size, the performance of FT, InContextLearning, and {\ModelName} is improved on both MRPC and RCT, which is consistent with the assumption that more data brings sufficient training.
Compared with baseline models, our model achieved consistent improvement over ManualPrompt and InContextLearning, verifying its effectiveness and scalability under different data sizes.

\paragraph{Effects of Prompt Length}
It is known that prompt-based methods are sensitive to many aspects of prompts, including contexts~\citep{jiang2020can}, orders~\citep{lu2021fantastically} and lengths~\citep{lester2021power}, and inappropriately designed prompts lead to bad performance. 
Here we study the effects of different prompt lengths on MRPC and RCT.
Considering the maximum number of tokens is 2048 for the input of GPT-3 API, too many prompt tokens (e.g., more than 100) cause additional costs and even failure of queries.
Therefore, we conduct experiments with length = 10, 25, 50, and 75. 
The results are shown in the Figure \ref{fig:ablation_study} (c) and (d).
With the increase in prompt length, the performance increases at first and decreases when the prompt length reaches 75 in most cases. 
We conclude that the approximate best prompt length is 50, since a shorter prompt length limits the representation capacity while a longer prompt length might involve noises contained in the training data and is hard to optimize.

\paragraph{Effects of Training Budgets}
Training budgets are essential factors for efficient-tuning methods.
An efficient method is expected to achieve good performance with as few as training budgets.
We measure the budgets by the number of prediction API calls and report the performance of our models with different numbers of API calls in Figure~\ref{fig:ablation_study} (c) and (d).
It is observed that with the increase of API calls, {\ModelName} under different settings obtains performance gains because of sufficient training.
All settings could converge within 12, 000 API calls.
We also find that the $50$-prompt performs well at first but the gap between $75$-prompt and $50$-prompt narrows down when the training budget grows, and finally, $75$-prompt achieves competitive performance.
It told us that if we do not have a sufficient training budget, it would be better to apply fewer prompt tokens, which are easier to optimize.

\begin{figure}[t]
\begin{center}
\includegraphics[scale=0.48, trim=0 0 0 0,clip ]{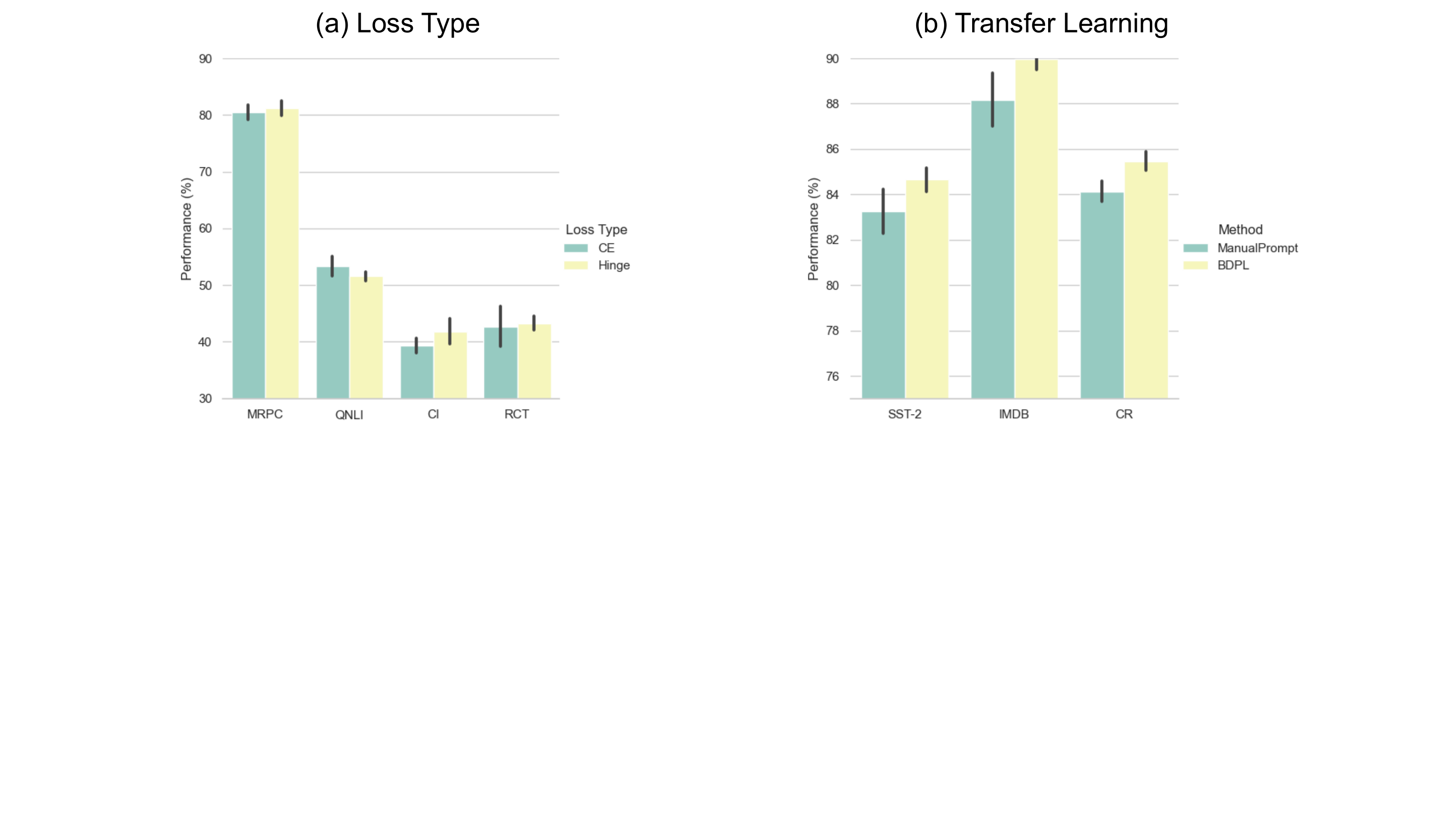}
\vskip -13 em
\caption{\textbf{(a) Ablations of loss function.} CE and Hinge represent cross-entropy loss and hinge loss, respectively. 
\textbf{(b) Transfer learning performance.} SST-2 is the source task while IMDB and CR are two tasks in the target domain.
}
\vskip -2 em
\label{fig:loss_type_transfer_learning}
\end{center}
\end{figure}

\paragraph{Effects of Different Objectives} 
\label{Effects of Different Loss Types}
In the previous experiments, the prompts are optimized with the cross-entropy loss, and here we explore the effectiveness of our model with different objectives.
With the same setting as our main experiment, we conduct further experiments with hinge loss on four datasets: MRPC, QNLI, CI, and RCT.
We find that our model with both objectives can achieve comparable results.
As shown in Figure \ref{fig:loss_type_transfer_learning} (a), the model with hinge loss outperforms that with cross-entropy loss on MRPC, CI, and RCT, but underperforms it on QNLI.
On average, our approach with hinge loss works as well as that with cross-entropy loss.
It is flexible enough to work with different objectives, and we hope to extrapolate to any kind of human-designed objectives.

\paragraph{Effects of Transfer Learning}
\label{Transfer Learning}
A critical advantage of discrete prompts is the possibility of transferring prompts learned from one task to another because discrete tokens share the same text space instead of specific latent space for continuous prompts. 
To verify the transferability of {\blackbox} optimized prompt tokens, we conduct experiments on three sentiment analysis datasets (\textit{i.e.}, SST-2~\citep{socher2013recursive}, IMDB~\citep{imdb}, and CR~\citep{hu2004mining}) with GPT-3-Babbage model in the 16-shot setting.
First, we use SST-2 as the source task following~\citet{vu2021spot} and optimize the discrete prompt tokens with our proposed {\ModelName}.
Then we obtain those selected prompt tokens and simply prepend them to the beginning of the input of the target task, IMDB and CR.
Our setting assumes no training data in the target domain, so we directly test the performance in the target task.
Following~\citet{wang2021entailment}, for CR, we randomly sample 2,000 instances as the test set.
The results are shown in Figure \ref{fig:loss_type_transfer_learning} (b).
Consistent with the previous observation in Section \ref{sec:performance}, {\ModelName} outperforms ManualPrompt in the source task by a large margin.
Moreover, learned prompts are helpful in two target tasks, demonstrating that our {\blackbox} method is robust under transfer learning settings.
The experimental results display the expansion potential of \textit{prompt transfer}, which is a promising practical application of {\ModelName}, especially when there are $N$ edge devices sharing a similar task, but they have no training data.
We can update the {\blackbox} prompts in a general domain and then transfer them to the target domain.

\begin{figure*}[t]
\centering
\includegraphics[scale=0.48, trim=0 10 0 0,clip]{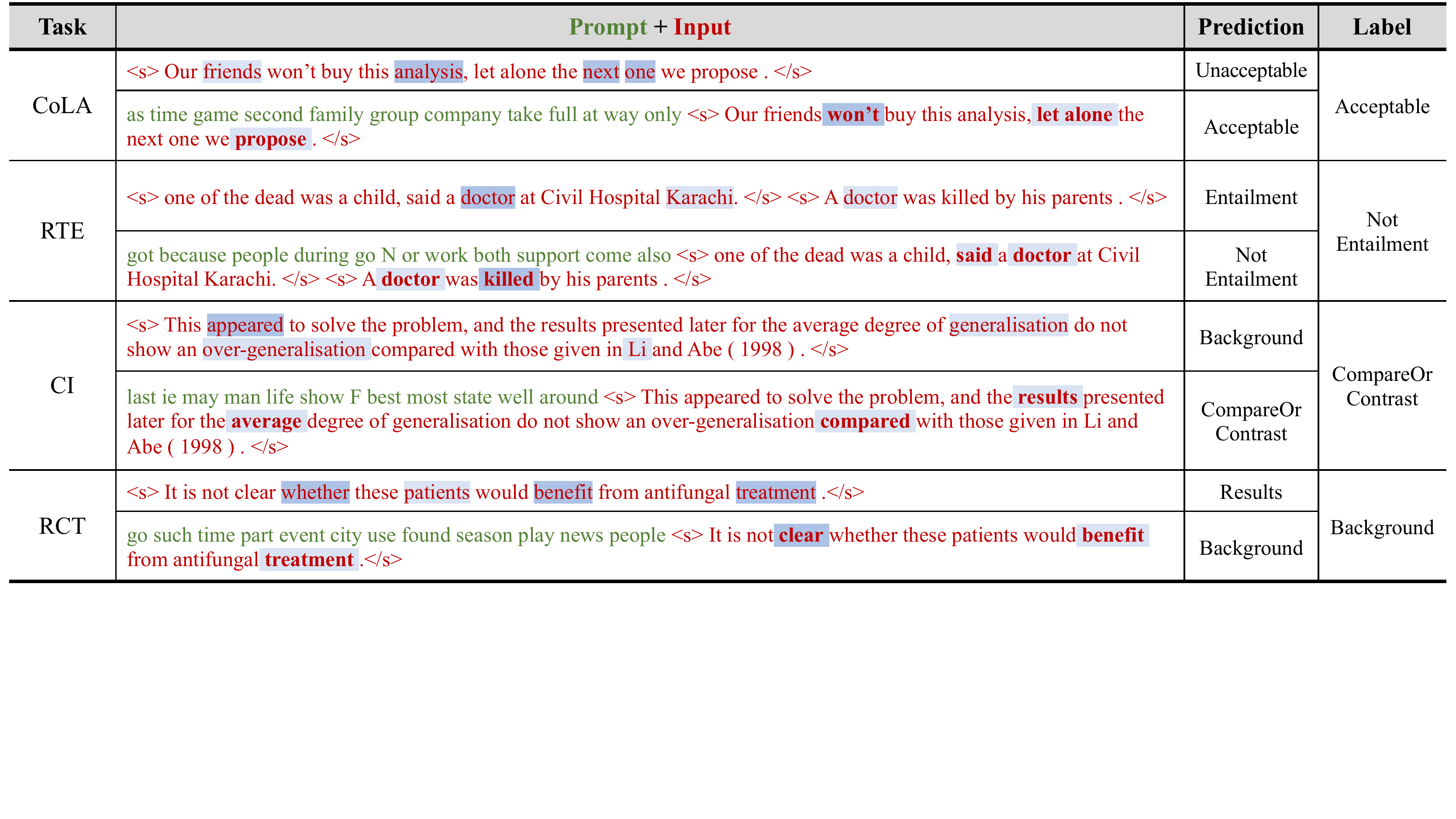} 
\vskip -8 em
\caption{
Four correctly predicted examples by {\ModelName}.
We display the prompts and salience map of the token \texttt{<s>}.
The prompts are in \textcolor{my_green}{green} and the input tokens are in \textcolor{my_red}{red}.
The salient tokens are highlighted in a blue background, where the darker color denotes the more dominant weights for the prediction.
}
\label{fig:attentions}
\vskip -2 em
\end{figure*} 

\subsection{Prompt Explanation}
To intuitively understand the prompts, we visualize the salience maps using the Language Interpretability Tool (LIT)~\citep{tenney2020language}.
We choose \textsc{CoLA}, RTE, CI, and RCT datasets because the sentences contained in these datasets are easy for human interpretation. 
The comparisons between models with discrete prompts and without them are shown in Figure~\ref{fig:attentions}.
By adding discrete prompt tokens, the model is able to find coherence-related clues. 
For example, \textsc{CoLA} aims to distinguish the acceptability of sentences.
The grammar-related word `won't' and phrase `let alone' dominate its prediction, which is consistent with the decision process of human beings.
Similar observations are found in other datasets.
Due to space limitations, more visualized
examples are not shown here.
Based on considerable empirical evidence, we conclude that {\ModelName} can capture helpful information to guide the model.

We notice that most of the optimized prompts are readable but in-comprehensible, useful to model improvement, but semantically confusing to humans. \citet{ilyas2019adversarial} find that neural networks rely on some \textit{non-robust} features to achieve the highest possible accuracy.
These \textit{non-robust} features are usually semantically meaningless to humans, but we can still train a well-performed model only using these features. 
We argue that while the optimized prompts that our method finds are meaningless to humans, they are useful for the models to make a more accurate prediction. 
In contrast, forcing the prompts to have semantic meaning may remove the useful information and leads to degraded performance. 
This observation is consistent with previous discrete prompt learning studies~\citep{shin2020autoprompt}.

\section{Related Work}
In this section, we present the review on prompt learning for pre-trained language models and black-box optimization.

\subsection{Prompts for Pre-trained Models}
Large pre-trained language models are of great importance and a standard paradigm is pre-training a language model on a large unlabeled corpus and then fine-tuning the pre-trained model on different supervised tasks.
This approach shows great improvements on lots of downstream tasks but it needs lots of computational resources to change all the parameters and has to save a copy for each task.
Therefore, prompt-based learning, which does not require tuning the large model, is proposed to solve the problem.
Based on the format of prompts, the prompt-based learning can be categorized into two kinds: discrete prompt~\citep{wallace2019universal, shin2020autoprompt, jiang2020can, gao2020making, ben2021pada} and continuous prompt~\citep{zhong2021factual, qin2021learning, hambardzumyan2021warp, liu2021gpt, han2021ptr,li2021prefix}. 
The discrete prompt is usually a sequence of tokens or natural language phrases while the continuous prompt is designed as a sequence of vectors.
However, all of these studies are limited to a {\whitebox} setting, which requires accessing all the parameters of a pre-trained model so that the gradients could be back-propagated to optimize the prompts.
Recently, \citet{sun2022black} proposed black-box tuning methods but they are optimizing continuous prompts, which is impractical in real applications, because most of the commercial APIs do not accept continuous vectors as input.
Our method, {\blackbox} prompt learning, provides a truly black-box solution with discrete optimization, which optimizes a set of discrete prompts without accessing the pre-trained model.
\highlight{There are some concurrent works \citep{deng2022rlprompt,hou2022promptboosting} exploring black-box tuning methods for large language models. PromptBoosting~\citep{hou2022promptboosting} ensembles a large number of weak learners by the AdaBoost algorithm to pair pre-generated prompts with different elements of the LM’s output distribution. 
They learn the verbalizer instead of discrete prompts of the language model, which is complementary to {\ModelName}. RLPrompt~\citep{deng2022rlprompt} generates discrete prompts by a policy network and optimizes it by a reward function.  In contrast, we apply a variance-reduced policy gradient estimator to optimize a few independent categorical probabilities to select the appropriate prompt tokens. 
In addition, we are the first to show that the black-box prompt learning methods can generalize to real-world large language models like GPT-3.}

\subsection{Black-Box Optimization}
One of the applications of {\blackbox} optimization is the score-based {\blackbox} adversarial attack~\citep{ilyas2018black, ilyas2018prior, huang2019black, ACFH2020square, cheng2019improving}, where the models are also invisible to the attacker. 
These studies use zeroth-order optimization methods such as natural evolution strategy (NES)~\citep{wierstra2014natural} to optimize the input and increase the loss to fool the model. 
Instead of deteriorating the models' performance in the adversarial attack, our direction is to find the inputs that improve the accuracy. Policy gradient \citep{sutton1999policy}, which also belongs to the black-box optimization, is widely used in reinforcement learning to find the best policy. In contrast to NES that can only be used to search in the continuous search, policy gradient allows the choice of discrete policy and can be used to find the optimal discrete prompts. 
{\ModelName} uses {\blackbox} optimization methods to find the optimal prompts, which is a novel application direction of these methods. 

\highlight{
Another line of research to adapt the black-box model is knowledge distillation (KD) \citep{hinton2015distilling}, which learns a student model with the outputs of large models. 
KD can be used to learn black-box models \citep{nguyen2022black, wang2021zero}, perform domain adaptation \citep{liang2022dine}, and adversarially attack the model \cite{zhang2022towards}. 
Despite the wide applications of black-box KD, learning the models with KD still requires a large number of queries and data for training the student network. 
On the contrary, our proposed approach, {\ModelName}, is much more lightweight, and only needs to train a few prompts with a small amount of data. 
Moreover, under the scenario of cloud-device collaboration, {\ModelName} only needs negligible computation on the edge devices while KD has to train the local student network with large computational resources.
Therefore, {\ModelName} is more practical for the scenario of cloud-device collaboration. 
}

\section{Conclusion}
This paper proposes a novel setting for text categorization namely {\blackbox} prompt learning, where a large pre-trained model is invisible so that the gradients cannot be back-propagated to update the prompts.
Compared with the standard pre-training then fine-tuning paradigm, our approach only requires updating very few parameters.
Compared with previous prompt-based methods, our approach does not require the visibility of pre-trained models, and thus it provides more flexibility in practical applications. 
We propose a {\blackbox} prompt learning method, \ModelName, which employs a variance-reduced policy gradient estimator to approximate the gradients, and then update the prompts.
Experimental results demonstrate that our approach outperforms all black-box methods and is comparable with white-box methods, illustrating the effectiveness of {\blackbox} optimization.
Experiments on the transfer learning settings further show the potential of our approach in realistic scenarios, where the pre-trained model is deployed on the cloud, and the prompt learning can be implemented on each device. 

In the future, we would like to explore the effectiveness of our proposed methods on more commercial classifiers, such as Google Cloud APIs, Microsoft Azure APIs and so on.
The black-box prompt learning for large multi-modal models~\citep{wang2021vlmo,singh2021flava,wang2021simvlm,zhou2022vlue,wang2022unifying,diao2023write} is another important scenario to explore in future work.

\section*{Acknowledgments}
We thank the anonymous reviewers for their valuable suggestions.
This work was supported by the General Research Fund (GRF) of Hong Kong (No. 16310222 and No. 16201320). 
Shizhe Diao, Ruijia Xu, and Yong Lin were supported by the Hong Kong Ph.D. Fellowship Scheme (HKPFS).

\bibliography{main}
\bibliographystyle{tmlr}

\newpage
\appendix
\section{Implementation Details}
\label{sec:appendix-implementation}

\subsection{Computing Infrastructure}
For experiments on GPT-3, we directly call its APIs without any GPU for computation.
For experiments on RoBERTa, they are conducted with NVIDIA 2080Ti GPUs with 11GB memory.

\subsection{Evaluation Measures}
For tasks from the GLUE Benchmark, we adopt Matthews correlation coefficient for \textsc{CoLA}, F1 for \textsc{MRPC}, and accuracy for \textsc{RTE} and \textsc{QNLI} following their original metric choices. 
We adopt macro-F1 for \textsc{CitationIntent}, \textsc{SciERC},  \textsc{RCT}, and \textsc{HyperPartisan} as evaluation metrics. 

\subsection{Bounds of Hyper-parameters}
\label{appendix:hyper-params}
\begin{table}[H]
  \centering
    \begin{tabular}{c|cc}
        \toprule
        \textbf{Hyper-parameter} & \textbf{GPT-3} & \textbf{RoBERTa} \\
        \midrule
        number of epochs & 30 & 30 \\
        train batch size & 4 & 32 \\
        eval and test batch size & 4 & 16 \\
        prompt length & \multicolumn{2}{c}{\{10, 12, 25, 50, 75\}} \\
        learning rate & \multicolumn{2}{c}{[1e-5, 1e-3]} \\
        dropout & \multicolumn{2}{c}{0.1} \\
        learning rate optimizer & \multicolumn{2}{c}{AdamW} \\
        loss type & \multicolumn{2}{c}{\{hinge loss, cross-entropy loss\}} \\
        \bottomrule
    \end{tabular}
     \caption{Bounds of hyper-parameters.}
     \label{tab:params}
\end{table}

\subsection{Configuration of Best Model}
The configuration of the best model for each dataset is shown in Table~\ref{tab:configuration}.

\begin{table*}[ht]
\centering
\begin{tabular}{lccccccccccc}
\hline
\textsc{Dataset} & MNLI & QQP & SST-2 & \textsc{MRPC} & \textsc{CoLA} & \textsc{QNLI} & \textsc{RTE}  & \textsc{CI} & \textsc{SE} & \textsc{RCT} & \textsc{HP} \\
\hline
prompt length    & 10 & 25 & 50 & 50    & 50  & 50  & 50  & 50 & 50 & 50 & 50 \\
learning rate  & 2e-4 & 1e-4 & 2e-4 & 1e-4  &  3e-4  &  2e-4  & 1e-4   & 1e-4 &  1e-4  & 1e-4 & 1e-4 \\
\hline
\end{tabular}
\caption{Configuration of the best model for each dataset. 
The rest hyper-parameters are the default value in Table \ref{tab:params}. 
}
\label{tab:configuration}
\end{table*}

\subsection{Manual Templates}
\label{appendix:manual_templates}
The manual templates are shown in Table~\ref{table:label_prompt}.

\begin{table*}[t]
\small
\centering
\renewcommand{\arraystretch}{1.1}
\begin{tabular}{cc}
\toprule
\textbf{Dataset}& \textbf{Template} \\\midrule

MNLI & sentence$_1$ entailment?\texttt{[MASK]}, sentence$_2$. (yes/no) \\
QQP & sentence$_1$ ?\texttt{[MASK]}, sentence$_2$. (yes/no) \\
SST-2 & sentence$_1$. It was \texttt{[MASK]}. (great/terrible) \\
MRPC & sentence$_1$ ?\texttt{[MASK]}, sentence$_2$. (yes/no) \\
CoLA & sentence$_1$. correct? \texttt{[MASK]}. (yes/no) \\
QNLI & sentence$_1$ entailment?\texttt{[MASK]}, sentence$_2$. (yes/no) \\
RTE & sentence$_1$ entailment?\texttt{[MASK]}, sentence$_2$. (yes/no) \\
CI & sentence$_1$. What is the intent?\texttt{[MASK]}. (background/compare/extends/future/motivation/uses) \\
SE & sentence$_1$. What is the relation?\texttt{[MASK]}. (compare/conjunction/evaluate/feature/ hyponym/part/used) \\
RCT & sentence$_1$. It is \texttt{[MASK]}. (background/conclusion/method/objective/result) \\
HP & sentence$_1$. It is \texttt{[MASK]}. (yes/no) \\
IMDB & sentence$_1$. It was \texttt{[MASK]}. (great/terrible) \\
CR & sentence$_1$. It was \texttt{[MASK]}. (great/terrible) \\
\bottomrule
\end{tabular}
\caption{Prompts and label descriptions of ManualPrompt method. Most of them are from ~\citet{gao2020making}. }
\label{table:label_prompt}
\end{table*}

\section{Projection Calculation} \label{sec:proj}

The projection from $\bs{z}$ to $\mathcal{C}$ can be calculated by:

\begin{algorithm}[htb!]
\caption{Projection from $\bs{z}$ to $\mathcal{C}$}
\label{alg:proj}
\begin{algorithmic}[1]
\REQUIRE a vector $\bs{z}$.
\STATE Solve $v_1$ from $\bs{1}^\top[\min (1, \max(0, \bs{z}-v_{1}^{*}\mathbf{1}))] - 1 = 0.$ \\
\STATE $\bs{p} \leftarrow \min (1, \max(0, \bs{p}-v_{1}^{*}\mathbf{1})).$ \\
\OUTPUT $\bs{p}$
\end{algorithmic}
\end{algorithm}

\begin{proof}
The projection from $\bs{z}$ to set $\mathcal{C}$ can be formulated in the following optimization problem:
\begin{align}
    &\min_{\bs{p}\in \mathbb{R}^n} \frac{1}{2}\|\bs{p}-\bs{z}\|^2,\nonumber\\
    s.t. & \mathbf{1}\top \bs{p}  = 1 \mbox{ and } 0\leq p_i \leq 1.\nonumber
\end{align}
Then we solve the problem with the Lagrangian multiplier method.
\begin{align}
    L(\bs{p},v) &= \frac{1}{2}\|\bs{p}-\bs{z}\|^2 + v(\mathbf{1}^\top \bs{p} -1)\\
    &=\frac{1}{2}\|\bs{p}-(\bs{z}-v\mathbf{1})\|^2 + v (\mathbf{1}^\top \bs{z}-1) -\frac{n}{2}v^2.
\end{align}
with $ 0 \leq p_i \leq 1$.
Minimize the problem with respect to $\bs{p}$, we have 
\begin{align}
    \tilde{\bs{p}} = \mathbf{1}_{\bs{z}-v\mathbf{1}\geq 1} + (\bs{z}-v\mathbf{1})_{1>\bs{z}-v\mathbf{1}>0}
\end{align}
Then we have
\begin{align}
g(v)=&L(\tilde{\bs{p}},v) \nonumber\\
   =& \frac{1}{2}\|[\bs{z}-v\mathbf{1}]_{-} + [\bs{z}-(v+1)\mathbf{1}]_{+}\|^2 \nonumber \\
   &+ v (\mathbf{1}^\top \bs{z}-1) -\frac{n}{2}v^2 \nonumber  \\
=&\frac{1}{2}\|[\bs{z}-v\mathbf{1}]_{-}\|^2 +\frac{1}{2}\|[\bs{z}-(v+1)\mathbf{1}]_{+}\|^2\nonumber \\
&+ v (\mathbf{1}^\top \bs{z}-1) -\frac{n}{2}v^2. \nonumber \\
g'(v)=& \mathbf{1}^\top [v\mathbf{1}-\bs{z}]_{+} +\mathbf{1}^{\top} [(v+1)\mathbf{1}-\bs{z}]_{-}\nonumber \\
&+(1^T\bs{z}-1)-nv \nonumber \\
=&\mathbf{1}^\top\min (1, \max(0, \bs{z}-v\mathbf{1})) - 1.\nonumber
\end{align}
It is easy to verify that $g'(v)$ is a monotone decreasing function with respect to $v$ and we can use a bisection method solve the equation $g'(v) = 0$ with solution $v^*_1$. Finally we have 
\begin{align}
    \bs{p}^* =& \mathbf{1}_{\bs{z}-v_1^*\mathbf{1}\geq 1} + (\bs{z}-v_{1}^*\mathbf{1})_{1>\bs{z}-v_1^*\mathbf{1}>0}\\ 
    =&\min (1, \max(0, \bs{z}-v_{1}^{*}\mathbf{1})).
\end{align}
\end{proof}

\section{Effects of Prompt Positions}

\highlight{In our main experiments, we follow existing prompt-based learning studies~\citep{li2021prefix} to prepend some prompt tokens to the original sequence. 
We also investigate the effects of prompt positions.
First, we introduce two new baselines based on GPT-3-babbage: suffix-tuning (placing prompt tokens after the original sequence) and infix-tuning (placing prompt tokens in the middle of the sequence). 
The results are shown in~\ref{tab:ablation_prefix}.
From the results, we observed that prefix-tuning outperforms suffix-tuning and infix-tuning by a large margin. 
We attribute this performance drop to the position embedding of the learned prompts. Compared with infix-tuning and suffix-tuning, prefix-tuning keeps the prompt tokens at the same position, which is consistent during the learning process.
However, the infix-tuning and suffix-tuning require the prompts to have adapting ability to dynamic positions. 
In addition, putting the prompt tokens in the middle of the sequence may break the semantic meaning of the original sequence, leading to even worse results than ManualPrompt. 
These observations are consistent with the prefix-tuning~\citep{li2021prefix}, so we decide to adopt prefix-tuning in our main method.
}

\begin{table}[t]
\centering
\setlength{\tabcolsep}{1mm}{
\begin{tabular}{l|ccccccccccc|c}
\toprule
\textsc{Dataset} & MNLI & QQP & SST-2 & \textsc{MRPC} & \textsc{CoLA} & \textsc{QNLI} & \textsc{RTE}  & \textsc{CI} & \textsc{SE} & \textsc{RCT} & \textsc{HP} & \textsc{Avg.} \\
\midrule
 \multicolumn{13}{c}{\textit{GPT-3 Babbage}}
\\ \midrule
FT  & 40.7\tiny{0.5} & 46.2\tiny{1.4} & 87.4\tiny{1.5} & 66.4\tiny{1.7} & 0.3\tiny{0.1} & 50.9\tiny{0.2} & 52.3\tiny{1.0} & 5.2\tiny{0.4} & 4.1\tiny{1.0} & 61.1\tiny{5.2} & 33.3\tiny{1.3} & 40.7 \\
MP & 28.9\tiny{0.8} & 34.1\tiny{1.2} & 83.5\tiny{1.2} & 62.4\tiny{3.2} & 0.2\tiny{1.0} & 48.8\tiny{1.4} & 51.2\tiny{0.6} & 31.4\tiny{2.8} & 1.7\tiny{0.5} & 21.7\tiny{2.3} & 27.2\tiny{1.5} & 35.6  \\
{ICL} & 35.7\tiny{0.9} & 45.2\tiny{1.9} & 86.2\tiny{1.4} & 65.4\tiny{1.7} & 2.6\tiny{0.0} & 48.3\tiny{0.9} & 51.5\tiny{0.4} & 13.1\tiny{1.5} & 2.5\tiny{0.9} & 36.7\tiny{1.8} & 32.2\tiny{1.4} & 38.1 \\
\midrule
{\ModelName}  & 41.0\tiny{0.6} & 50.4\tiny{1.5} & 86.4\tiny{1.1} & 67.7\tiny{1.2} & 2.8\tiny{0.1} & 52.1\tiny{0.3} & 53.1\tiny{1.0} & 40.2\tiny{2.5} & 3.2\tiny{0.8} & 45.2\tiny{2.2} & 30.4\tiny{2.3} & 43.0 \\
{\ModelName}-infix  & 25.1\tiny{1.1} & 35.6\tiny{1.8} & 80.2\tiny{2.3} & 60.3\tiny{1.4} & 0.5\tiny{0.2} & 50.8\tiny{0.3} & 51.2\tiny{1.0} & 15.2\tiny{2.9} & 2.0\tiny{0.7} & 10.5\tiny{0.9} & 20.2\tiny{1.7} & 32.0  \\
{\ModelName}-suffix  & 39.1\tiny{1.7} & 47.5\tiny{2.1} & 85.6\tiny{3.1} & 64.2\tiny{2.6} & 1.1\tiny{0.4} & 51.8\tiny{1.6} & 52.9\tiny{1.2} & 23.7\tiny{1.6} & 2.9\tiny{0.7} & 43.1\tiny{2.3} & 28.8\tiny{1.2} & 40.1  \\
\bottomrule
\end{tabular}
}
\caption{
The effects of different prompt token positions.
We report average scores across three random seeds, with standard deviations as subscripts.
\textsc{Avg.} denotes the average score across all tasks.
FT: GPT-3's FineTuning. 
MP: ManualPrompt. 
ICL: InContextLearning. 
}
\label{tab:ablation_prefix}
\end{table}

\section{Case Studies}
\begin{figure*}[t]
\centering
\includegraphics[scale=0.48, trim=0 0 0 0,clip ]{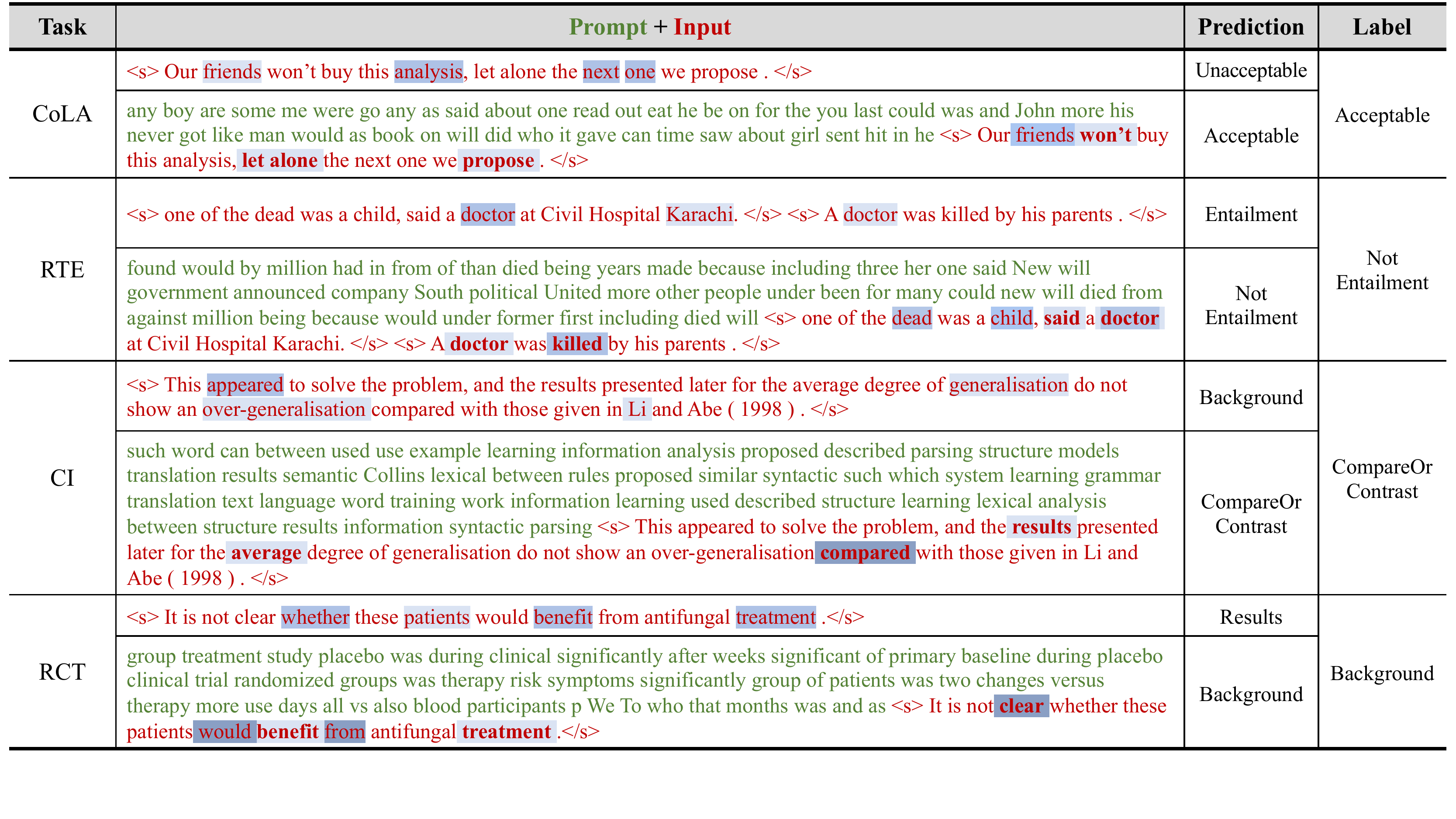} 
\vskip -3 em
\caption{
Four correctly predicted examples by {\ModelName}.
We display the prompts and salience map of the token \texttt{<s>}.
The prompts are in \textcolor{my_green}{green} and the input tokens are in \textcolor{my_red}{red}.
The salient tokens are highlighted in a blue background, where the darker color denotes the more dominant weights for the prediction.
}
\label{fig:attentions_appendix}
\vskip -1 em
\end{figure*}

\highlight{More case studies for prompt explanation are provided in Figure~\ref{fig:attentions_appendix}.
}

\section{Performance with More Data}
\label{appendix:overfitting}

\begin{figure}[t]
\begin{center}
\includegraphics[scale=0.48, trim=0 0 0 0,clip ]{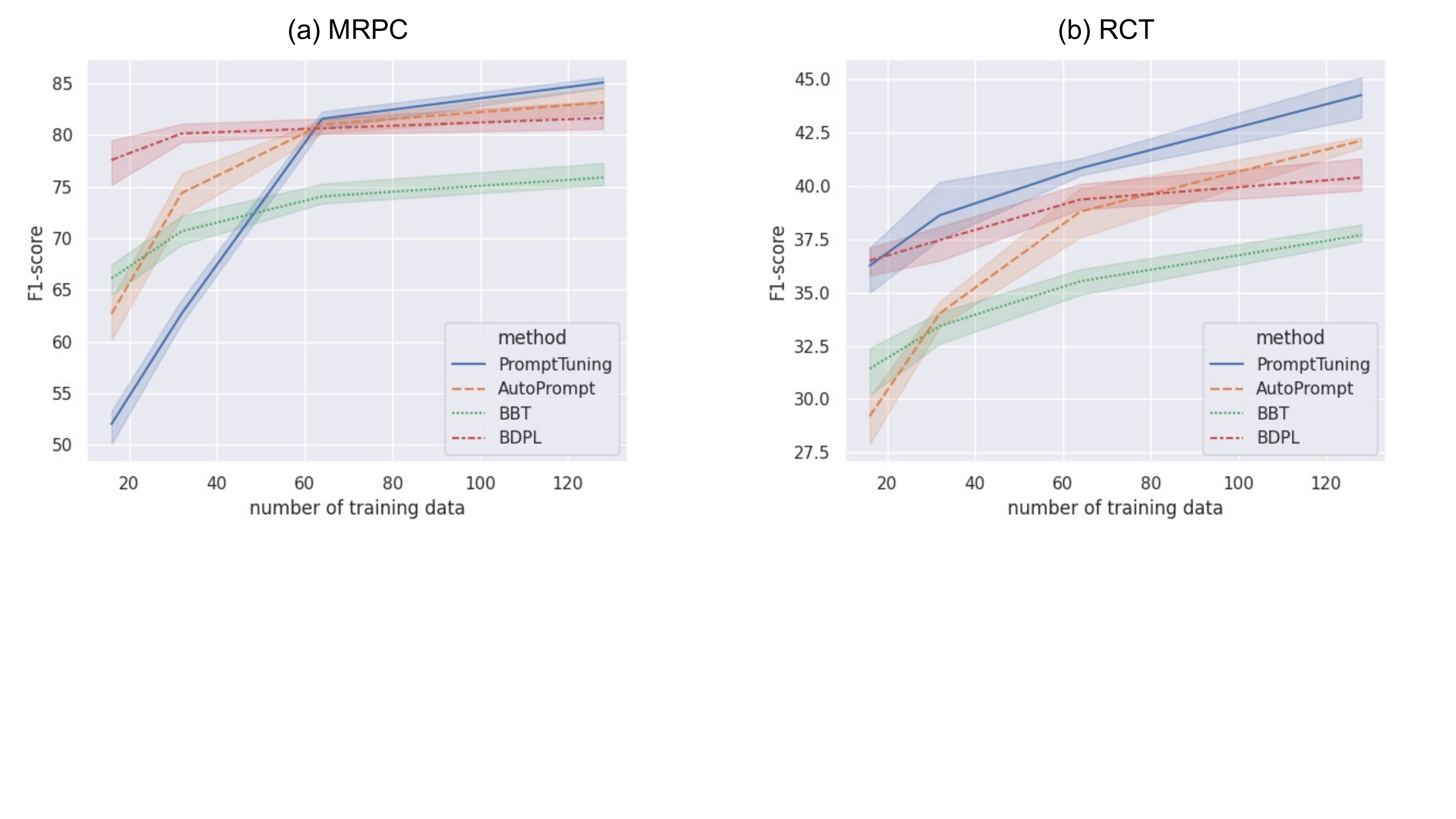}
\vskip -9 em
\caption{The effects of training data size on MRPC and RCT with RoBERTa-large model.
PromptTuning and AutoPrompt are two white-box methods. BBT and BDPL are two black-box methods.
}
\label{fig:whitebox_overfit}
\end{center}
\end{figure}

\highlight{
It is not very straightforward that black-box methods can outperform the white–box methods (i.e., PromptTuning and AutoPrompt). 
However, some similar observations were reported in previous and concurrent black-box studies. 
For example, BBT outperforms PromptTuning and AutoPrompt, and RLPrompt outperforms AutoPrompt.
According to our experiments, we attribute this phenomenon to the overfitting of white-box methods in terms of the given few-shot examples. 
First, in our experiments, we found that PromptTuning and AutoPrompt have lower training losses. 
It suggests that white-box methods tend to overfit the small training data (our experiments are 16-shot). 
Furthermore, we gradually increase the number of training data and verify whether the experimental results are consistent with our conjecture. 
Specifically, we increase the training data from 16-shot to 32-shot, 64-shot, and 128-shot. 
The results are shown in Figure~\ref{fig:whitebox_overfit}. 
It is observed that as the amount of training data increases, the white-box methods will outperform black-box methods, which demonstrates that white-box methods are better than black-box methods given sufficient data.
Based on these experiments, we believe black-box methods are good at few-shot settings due to their exploration mechanism, which can mitigate overfitting.
}

\end{document}